\documentclass[10pt,twocolumn,letterpaper]{article}

\usepackage{pdfpages}
\usepackage{cvpr}
\usepackage{times}
\usepackage{epsfig}
\usepackage{graphicx}
\usepackage{amsmath}
\usepackage{amssymb}
\usepackage{comment}
\usepackage{subcaption}
\usepackage[breaklinks=true,bookmarks=false]{hyperref}
\cvprfinalcopy 

\def\cvprPaperID{1} 

\title{ Empirically Analyzing the Effect of Dataset Biases \\on Deep Face Recognition Systems}
\author{
\begin{tabular}[t]{c c c c c c} 
Adam Kortylewski  & Bernhard Egger & Andreas Schneider & \\Thomas Gerig & Andreas Morel-Forster & Thomas Vetter\\
\end{tabular}\vspace{.1cm}\\
Department of Mathematics and Computer Science\\
University of Basel\\
}
\begin{document}
\maketitle

\begin{abstract}
It is unknown what kind of biases modern in the wild face datasets have because of their lack of annotation.
A direct consequence of this is that total recognition rates alone only provide limited insight about the generalization ability of a Deep Convolutional Neural Networks (DCNNs). We propose to empirically study the effect of different types of dataset biases on the generalization ability of DCNNs.
Using synthetically generated face images, we study the face recognition rate as a function of interpretable parameters such as face pose and light.
The proposed method allows valuable details about the generalization performance of different DCNN architectures to be observed and compared. 
In our experiments, we find that:
1) Indeed, dataset bias has a significant influence on the generalization performance of DCNNs.
2) DCNNs can generalize surprisingly well to unseen illumination conditions and large sampling gaps in the pose variation. 
3) Using the presented methodology we reveal that the VGG-16 architecture outperforms the AlexNet architecture at face recognition tasks because it can much better generalize to unseen face poses, although it has significantly more parameters.
4) We uncover a main limitation of current DCNN architectures, which is the difficulty to generalize when different identities to not share the same pose variation. 
5) We demonstrate that our findings on synthetic data also apply when learning from real-world data.
Our face image generator is publicly available to enable the community to benchmark other DCNN architectures.
\end{abstract}

\section{Introduction}
Deep face recognition systems \cite{taigman2014deepface,schroff2015facenet,Parkhi15} have achieved remarkable performances on large scale face recognition datasets such as Labeled Faces in the Wild \cite{huang2007labeled} or Megaface \cite{kemelmacher2016megaface} in the recent years. However, the precise limitations of face recognition systems is unclear, since a fine-grained annotation of nuisance transformations, such as the face pose or the illumination conditions is practically unfeasible on such large scale datasets. In addition, this lack of annotation makes it difficult to analyze if certain limitations are caused by properties of a particular DCNN architecture or simply by a bias in the data.
\begin{figure}[t]
    \centering
    \includegraphics[width=.35\textwidth]{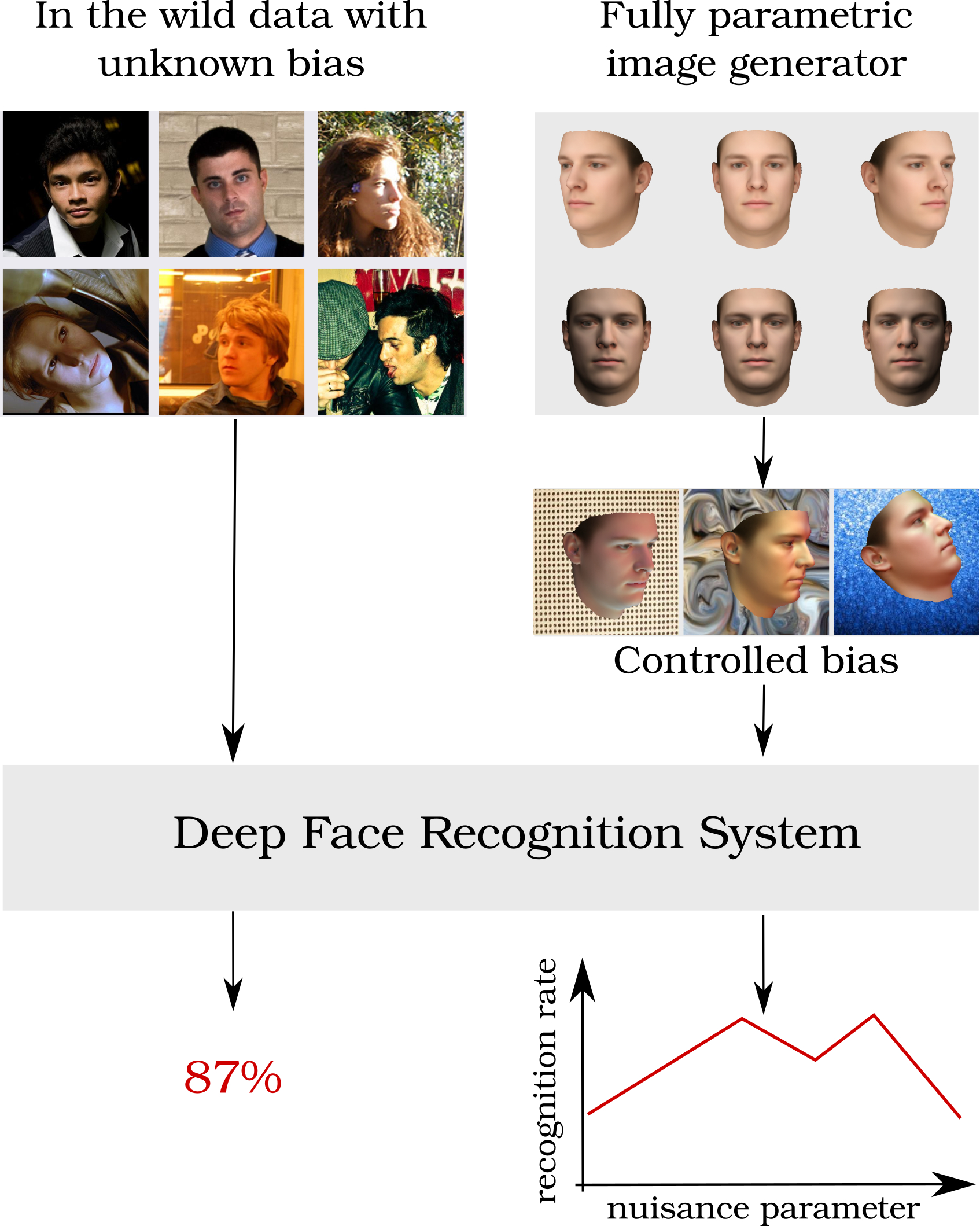}
    \caption{Importance of annotated datasets for diagnosing deep face recognition systems. Left: In the wild data does not permit any analysis of the generalization ability that goes beyond the total recognition rate. Right: Our proposed synthetic face image generator enables a detailed analysis of the recognition score as a function of the most relevant nuisance transformations, such as the face pose, illumination conditions, facial expressions and dataset bias.
    }
    \label{fig:frontpage}
\end{figure}

We propose to overcome this lack of transparency by evaluating face recognition systems on synthetic face images that are generated with a parametric 3D Morphable Face Model \cite{blanz1999morphable}. In particular, we introduce a face image generator that can create ground-truth face recognition datasets with a fine-grained control over parameters that define the facial identity, such as shape and texture, but also over nuisance parameters, such as light, camera and face pose (Figure~\ref{fig:frontpage}). We propose to make use of these fully annotated datasets for the empirical analysis of common DCNN architectures at the task of face recognition on a common ground. Our main contributions are:
\begin{itemize}
    \item A fully parametric face image generator based on a 3D Morphable Face Model that synthesizes naturally looking face images with precise annotation of the main sources of image variation. Our face image generator is publicly available. 
    \item A methodology for the systematic empirical analysis of DCNN architectures at the task of face recognition. Thereby, we introduce different kinds of biases in the training data and compare the generalization performance of different DCNN architectures on unbiased test data.
    \item We find several interesting properties about the generalization ability of DCNNs at the task of face recognition, which we summarize in the following:
\end{itemize}
i) DCNNs can generalize surprisingly well to incoming light from previously unobserved directions, even if it induces strong changes of the facial appearance (Section~\ref{sec:experiments-bias}).
ii) It is well known that DCNNs with the VGG-16 architecture can generalize better than with the AlexNet architecture at face recognition tasks. Using the presented methodology we reveal that VGG-16 outperforms AlexNet, \textit{because} it can much better generalize to unseen face poses, although it has significantly more parameters. (Section~\ref{sec:experiments-bias}-\ref{sec:experiments-realdata}).
iii) If large variations of the yaw pose are not reflected in the training data, then DCNNs do not recognize faces in large yaw poses at test time (Section~\ref{sec:experiments-bias}).
iv) In a real world scenario, not all identities in the training data share the same pose and illumination settings. We simulate this setting and observe that DCNNs have major difficulties in generalizing when different identities do not share the same pose variation in the training data (Section~\ref{sec:experiments-disentanglement}).
v) When training DCNNs on real data we observe similar generalization patterns as on our synthetically generated data (Section~\ref{sec:experiments-realdata}). 
The paper is structured as follows: We discuss related work in Section~\ref{sec:related} and introduce our face image generator in Section~\ref{sec:generator}. We evaluate the generalization ability of different DCNN architectures under biased training data in Section~\ref{sec:experiments}.
We conclude our work and discuss caveats in Section~\ref{sec:conclusion}.
\begin{figure*}[ht]
    \includegraphics[width=\textwidth]{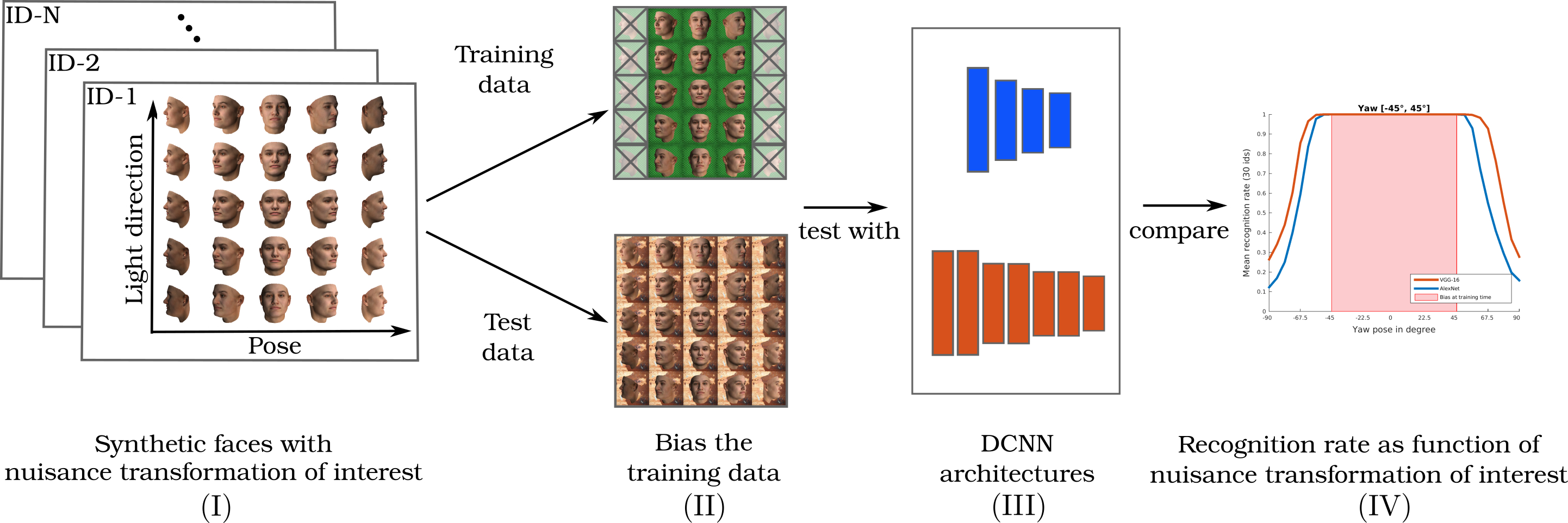}
   \caption{Experimental setup for our empirical analysis of the effect of biased training data on the generalization ability of different DCNN architectures. (I) We generate synthetic identities with a 3D Morphable Face Model and render them in different poses and illumination conditions. We simulate background variation by overlaying the faces on different textures. (II) We bias the training data by removing certain viewpoints from the training set. (III) We train common DCNN architectures on the biased training data. (IV) The annotation of the test data makes possible to analyze the recognition rate as a function of the face pose. It provides fine-grained information about the generalization ability of the different DCNN architectures.}
\label{fig:ExperimentOverview}
\end{figure*}
\section{Related Work}
\label{sec:related}
\noindent\textbf{Comparison of DCNN architectures.} 
Chatfield et al. \cite{chatfield2014return} compare different DCNN architectures on a common ground and found that deep architectures achieve superior performance to shallow architectures given extensive data augmentation. Mehdipour et al. \cite{mehdipour2016comprehensive} compare the VGG-face network \cite{Parkhi15} with the lightend CNN \cite{wu2015light} on several face datasets for which nuisance transformations such as pose variation or illumination changes were labeled. Their evaluation revealed that VGG-face achieves superior performance over the lightend CNN at most datasets. However, their diagnosis is limited by the fact that publicly available datasets only provide labels for a subset of all relevant nuisance transformations. In addition, pose transformations are mostly limited to changes in the yaw pose and are only sampled very roughly. The authors of \cite{reale2016analysis} evaluate several DCNNs at face recognition with respect to the influence of the size of the dataset as well as false labeling. However, it is difficult to interpret their results as they also have not taken into account the dependence between the different nuisance transformations. Karianakis et al. \cite{karianakis2016empirical} empirically study the influence of scale and location nuisances on the generalization ability of DCNNs at the task of object recognition and find that DCNNs can become invariant to these nuisances when learned from large datasets.\\
In this work, we study complex nuisance transformations such as 3D pose as well as illumination variations. In addition, we analyze the dependence between nuisance transformations and the effect of different sampling intervals of those transformations on the generalization performance. Furthermore, we evaluate the influence of biases in the sampling of nuisance transformations on the generalization performance of different DCNN architectures, such as e.g. biases to frontal face poses.\\
\textbf{Evaluation of Deep Learning theories.} Recently, theories have been developed to support the understanding of the internal mechanisms in deep learning systems in terms of symmetry regularization \cite{anselmi} and the information bottleneck \cite{tishby}. Especially for the task of image analysis, several approaches have been proposed to encode symmetries of data points w.r.t. transformations directly into the network structure, such as e.g. in Group Equivariant Networks \cite{cohen2016group}, Deep Symmetry Networks \cite{gens2014deep}, Transforming Autoencoders \cite{hinton2011transforming} or Capsule Networks \cite{sabour2017dynamic}. However, in order to evaluate the validity of these approaches it is of central importance to have full control over the transformation symmetries in realistic data. Our work in this paper enables such a detailed evaluation by providing full parametric control over variations in shape, pose, appearance and the illumination in face images.\\
\textbf{Diagnosis of computer vision with simulated data.} Synthetic datasets have been proposed for the evaluation of computer vision tasks such as optical flow \cite{butler2012naturalistic}, autonomous driving systems \cite{chen2015deepdriving}, object detection \cite{gupta2014learning}, pose estimation \cite{park2015articulated,ionescu2014human3} or for pre-training DCNNs \cite{gaidon2016virtual}. Qiu and Yuille \cite{qiu2017unrealcv} developed UnrealCV, a computer graphics engine for the diagnosis of computer vision algorithms at scene analysis. Their experiments reveal a large variation of the recognition performance of DCNNs at object detection across different viewpoints. In this paper, we take a similar approach to face recognition. In addition to leveraging computer graphics for face image generation, our data generator also enables the statistical variation of face shapes and textures which is learned from a population of 3D face scans. \\
\textbf{Face datasets with labeled nuisance transformations.} Several face databases are available with labeled nuisance transformations such as illumination variations in the CMU Multi-PIE \cite{gross2010multi} and Extended Yale \cite{georghiades2001few} databases or pose variations in the Color FERET \cite{phillips1998feret} database. However, theses datasets are of very small scale compared to modern in the wild databases and the sampling intervals along different transformations are coarse. Recently, Kemelmacher-Shlizerman et al. \cite{kemelmacher2016megaface} presented Megaface, a database with $690K$ identities and large scale pose annotations for in the wild faces. They demonstrate the importance of large amounts of "distractors", people who are not in the training set, on the performance of face recognition systems. However, the poses in Megaface are estimated from detected landmark positions, thus it is unclear how accurate these annotations are. Furthermore, the illumination conditions are not labeled and the number of training images per identity is rather small. Our simulation approach is complementary to current face recognition datasets, since it enables a fully controlled composition of training and test datasets. In particular, it makes possible to vary nuisance transformations in fine intervals, to arbitrarily scale the number of identities, as well as the number of training images per identity, in the training and test set. 

\section{Face Image Generator}
\label{sec:generator}
We propose to use a fully parametric generator for the synthesis of face images with detailed annotation of the most relevant nuisance transformations. Our generator is based on a 3D Morphable Model \cite{blanz1999morphable} of face shape, color and expression. In particular, we use the Basel Face Model 2017 (BFM-2017) \cite{gerig17bfm} which is learned from 200 neutral face scans and 160 expression deformations. The shape and color models are parametrized with 199 principal components each, the expressions are parametrized with 100 principal components. Natural looking, three dimensional faces with expressions can be generated by sampling from the statistical distribution of the model.\\
Using computer graphics we generate a 2D image from a 3D face, sampled from the model. We use a pinhole camera model as well as a spherical harmonics based illumination model \cite{ramamoorthi2001efficient, basri2003lambertian}. We represent the illumination as an environment map and approximate it with the first three bands of spherical harmonics, leading to 27 illumination parameters, 9 per color channel and use the prior introduced in \cite{egger2018occlusion}.
We use a non-parametric background model that chooses random background textures from the data provided in the describable texture database \cite{cimpoi14describing}.
The face image generator is built on the scalismo-faces software framework \cite{scalismo-faces}. The generator is publicly available \footnote{\url{https://github.com/unibas-gravis/parametric-face-image-generator}}.
The generator makes possible to generate infinite amount of face images with detailed labeling of the most relevant sources of image variation.
Example images synthesized from the generator are illustrated in Figure~\ref{fig:ExperimentOverview}. The fine-grained control over the data enables us to systematically analyze different DCNN architectures on a common ground at the task of face recognition in the next section. 
\section{Experiments}
\label{sec:experiments}
In this section, we demonstrate the importance of having fine-grained control over the image variation in the training and test dataset. 
In particular, it enables us to decompose the \textit{total recognition rate (TRR)} as a function along the axis of nuisance transformations. With this tool at hand, we study how biases in the training data, such as e.g. missing viewpoints of a face or unobserved illumination conditions, affect the generalization of DCNNs to unseen data at test time.\\
We describe the experimental setup in the following Section~\ref{sec:experiments-setup}. In Section~\ref{sec:experiments-bias}, we analyze the generalization performance of DCNNs if nuisance transformations are only partially observed at training time. In Section~\ref{sec:experiments-disentanglement}, we test the ability of DCNNs to disentangle image variations induced by nuisance transformations from identity changes. Section~\ref{sec:experiments-realdata} demonstrates that the generalization patterns we observe on synthetic data can also be observed when training on real data.
\subsection{Experimental Setup}
\label{sec:experiments-setup}
Figure~\ref{fig:ExperimentOverview} schematically illustrates our experimental setup. We generate synthetic images of different facial identities and transform them along the axes of the nuisance transformations that we want to study (Figure~\ref{fig:ExperimentOverview} $(I)$). In order to be able to study the influence of a particular bias in the training data, we must minimize the number of sources of nuisance transformations in the experiments. Therefore, we focus on varying the appearance of a face only in terms of the yaw pose as well as by rotating a directed light source around the face at a fixed inclination of $55^\circ$. We simulate strong background variations, which are common in real world data, by sampling random textures from our empirical background model. All other nuisance parameters are fixed. 
We illustrate samples of the face image generator with the nuisance transformations that we consider in our experiments in Figure~\ref{fig:ExperimentOverview}.
After splitting the synthetic data into a training and test set we bias the training data e.g. by removing certain face poses (Figure~\ref{fig:ExperimentOverview} $(II)$).
Subsequently, we train different DCNN architectures on the biased training data (Figure~\ref{fig:ExperimentOverview} $(III)$) and evaluate how well the DCNNs generalize to the unbiased test data. The fully parametric nature of the synthetic data, allows us to evaluate the recognition rate as a function of the biased nuisance transformation (Figure~\ref{fig:ExperimentOverview} $(IV)$).\\
In our experiments, we focus on comparing DCNNs with a significantly diverging performance at face recognition (AlexNet and VGG-16), as our methodology makes possible to study \textit{why} exactly one performs better than the other.
We test these networks at the task of face classification. The task is to recognize a face from an image, for which the identity is known at training time. Another common way of performing face recognition is to use the neural representation of the penultimate layer and to perform recognition via nearest neighbor in this feature space \cite{Parkhi15}. However, we focus on diagnosing the performance of DCNNs on the task that they were explicitly optimized on.\\
\textbf{Parameter Settings.} The size of the images is set to $227\times227$ pixels. We train the DCNNs with stochastic gradient descent (SGD) and backpropagation with the Caffe deep learning framework \cite{jia2014caffe} via the Nvidia DIGITS training system. Every DCNN is trained from scratch for $30$ epochs with a base learning rate of $l=0.001$ which is multiplied every $10$ epochs by $\gamma=0.1$. We use $L_2$ regularization with a weight regularization parameter of $\lambda=\frac{l}{100}$. If not stated otherwise, the data is uniformly sampled across the pose and illumination axes in the specified ranges. The training data consists of $30$ different identities, which we obtain by randomly sampling the shape and appearance parameter of the 3DMM. The images in the test set always reflect an unbiased sampling of the nuisance transformation that we want to study. For the yaw pose, we sample the parameter space at intervals of $\frac{\pi}{32}$ radian and for the direction of light at $\frac{\pi}{16}$ radian. Each face image is overlayed on $50$ different background textures in the training as well as in the test set. 

\subsection{Common bias over all facial identities}
\label{sec:experiments-bias}
\begin{figure}[b]
    \centering
    \subcaptionbox{\label{fig:SetupDisentanglement-partial}}{\includegraphics[scale=0.045]{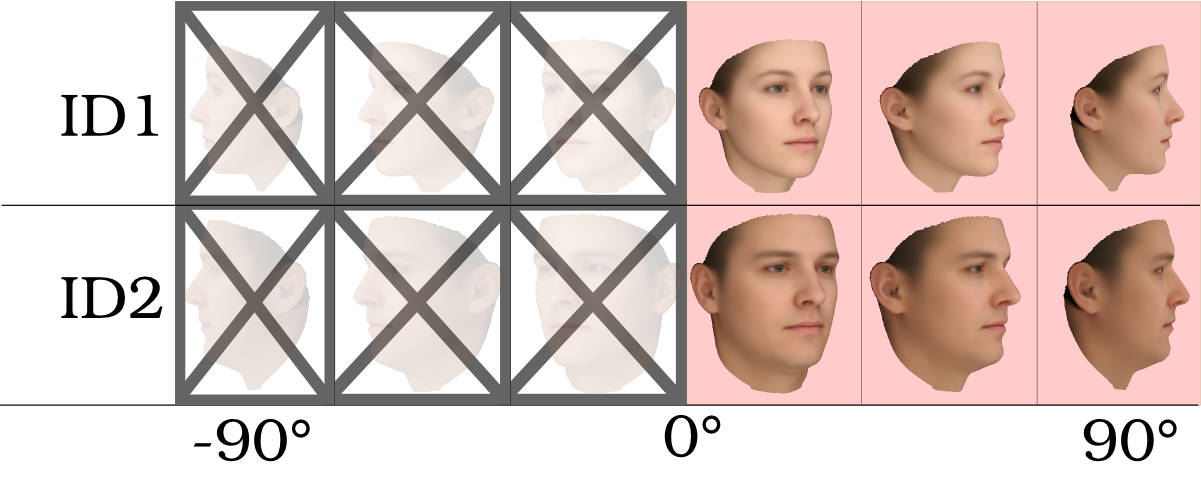}}
    \subcaptionbox{\label{fig:SetupDisentanglement-full}}{\includegraphics[scale=0.09]{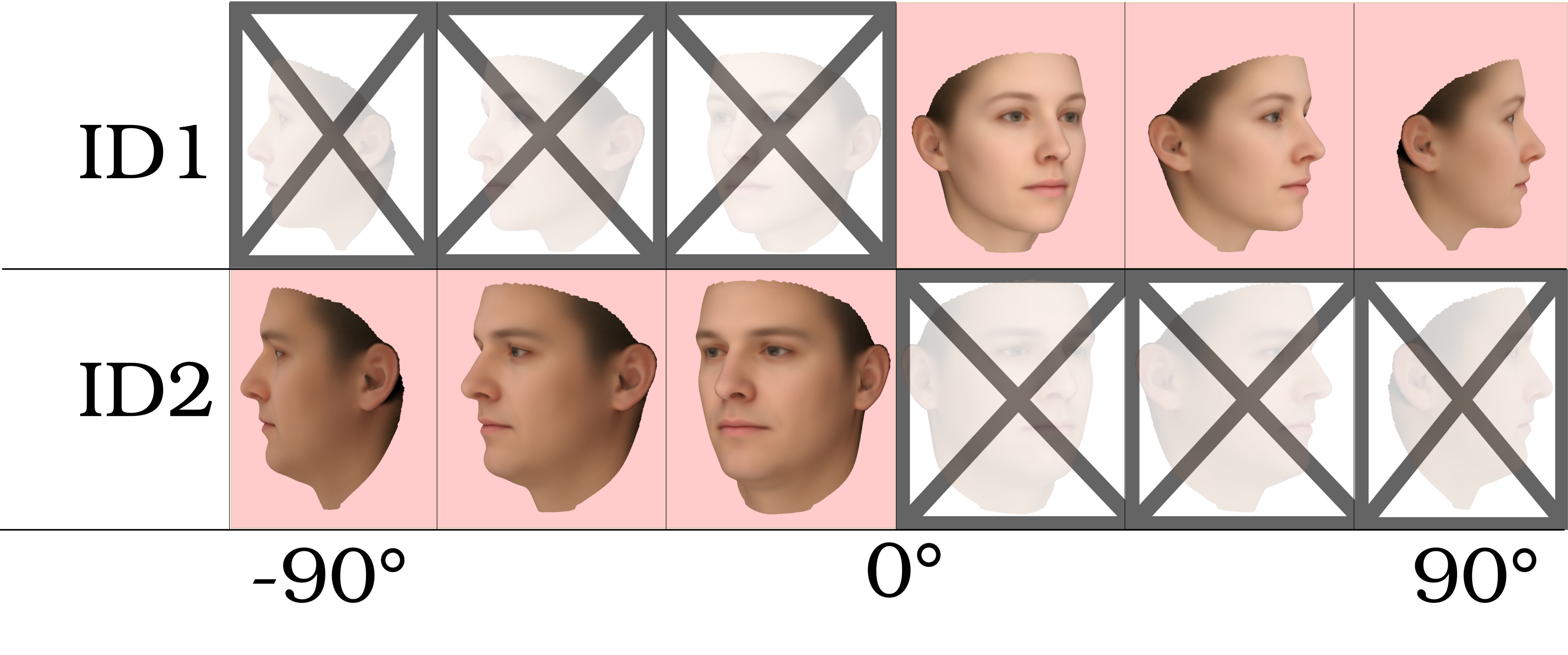}}
    \caption{Different types of biases illustrated on the example of yaw pose. Faces with red background are part of the training set. (a) The same bias is applied to all the identities in the training set. Thus, the pose variation space is only partially observed. We use this setup in Section~\ref{sec:experiments-bias} and \ref{sec:experiments-realdata}. (b) For each half of the identities an alternating half of the pose transformation is applied. Thus, the full pose transformation space is reflected in the data (Section~\ref{sec:experiments-disentanglement} \& \ref{sec:experiments-realdata}). }
    \label{fig:SetupDisentanglement}
\end{figure}

In this Section, we limit the range of nuisance transformations in the training data and analyze if DCNNs can generalize to the unobserved nuisance transformations. Furthermore, we analyze the effect of biasing the number of training examples to frontal poses. We apply the same bias to all identities in the training set (Figure~\ref{fig:SetupDisentanglement-partial}).\\
\begin{figure}
    \centering
    \subcaptionbox{\label{fig:DatasetBiasPlotDense-front}}{\includegraphics[scale=0.25]{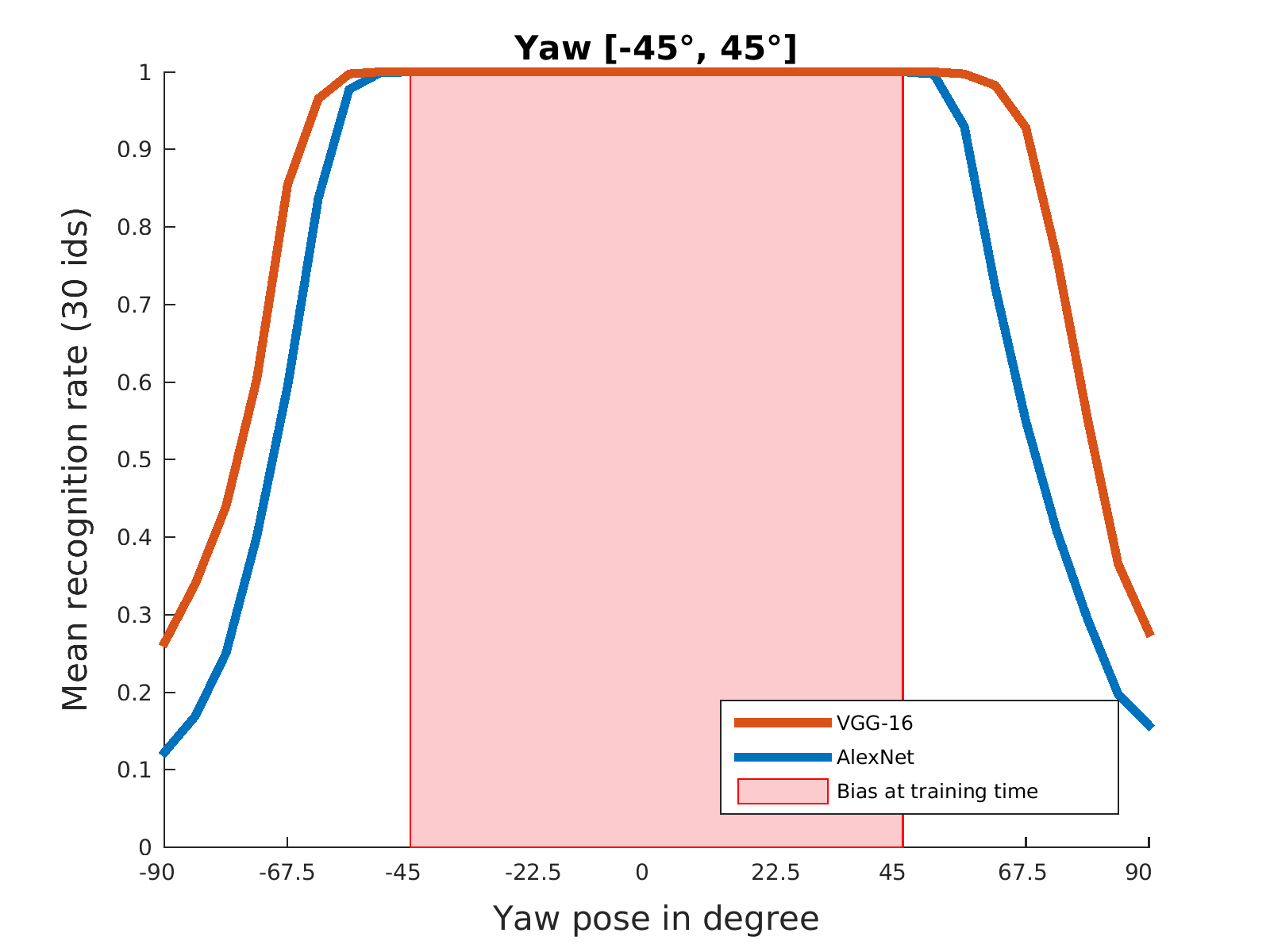}}
    \subcaptionbox{\label{fig:DatasetBiasPlotDense-left}}{\includegraphics[scale=0.25]{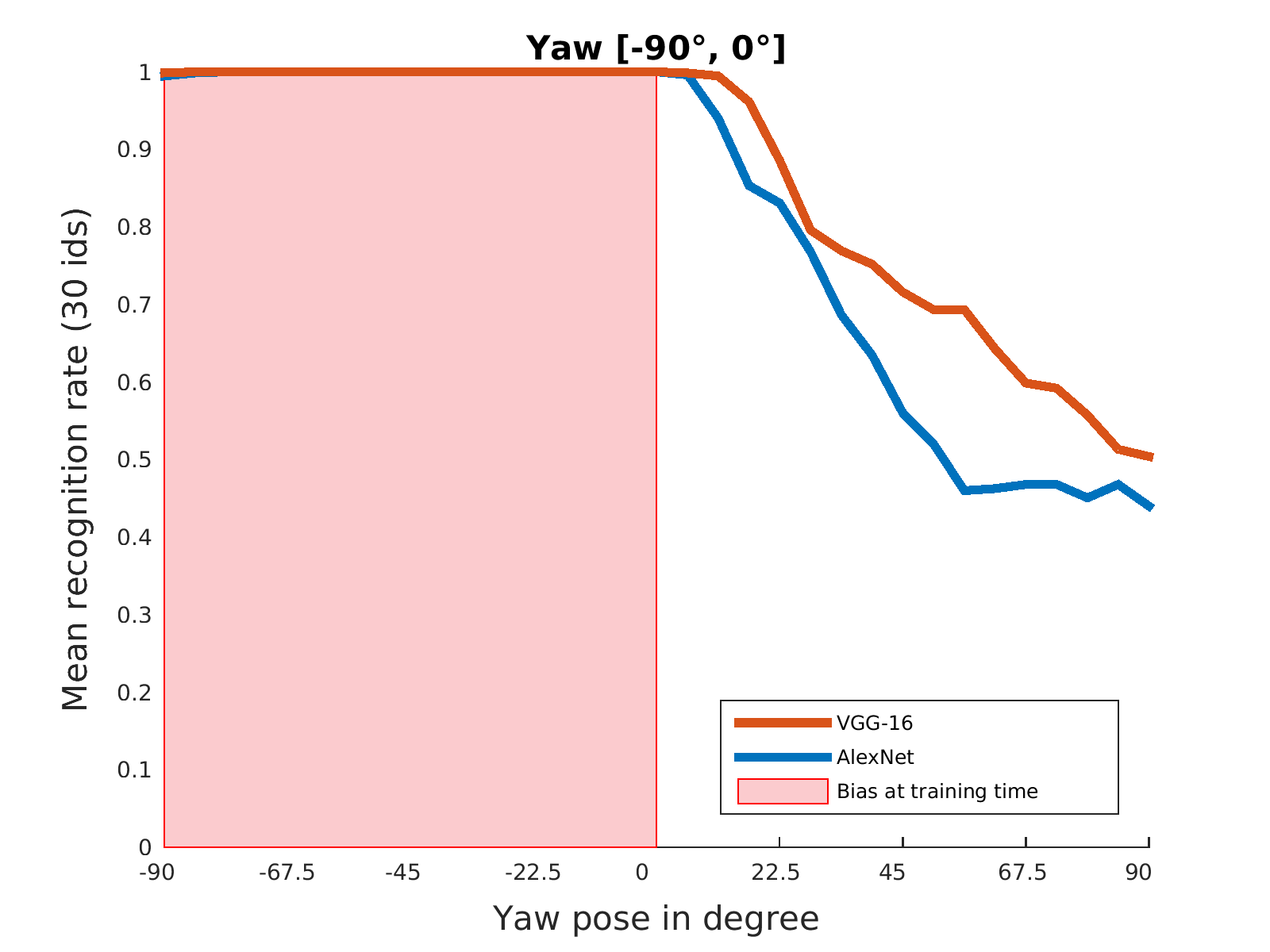}}
    \caption{Effect of restricting the range of yaw poses at training time. (a) Yaw pose restricted to the range $[-45^\circ,45^\circ]$. AlexNet TRR: $77.6\%$; VGG-16 TRR:$85.9\%$. (b) Yaw pose restricted to the range $[-90^\circ,0^\circ]$. AlexNet TRR: $81.8\%$; VGG-16 TRR:$86.9\%$. In both setups the DCNNs cannot recognize faces well from previously unobserved views. VGG-16 achieves a higher TRR due to the better generalization to large unseen yaw poses.}
    \label{fig:DatasetBiasPlotDense}
\end{figure}
\textbf{EXP-1: Bias in the range of the yaw pose.} In the following experiments, we limit the range of the yaw pose in the training data. The light direction is fixed to be frontal. Figure~\ref{fig:DatasetBiasPlotDense-front} illustrates the recognition performance as a function of the yaw pose, when faces in the training set are restricted to a yaw pose range of $[-45^\circ,45^\circ]$. Both DCNNs achieve high recognition rates for the observed yaw poses. However, the recognition performance drops significantly when faces are outside of the observed pose range. The same generalization pattern can be observed when restricting the faces at training time to a yaw pose range of $[-90^\circ,0^\circ]$ (Figure~\ref{fig:DatasetBiasPlotDense-left}). In both experiments, the VGG-16 network achieves higher overall recognition rates, \textit{because} it generalizes better to larger unseen yaw poses. 
\begin{figure}
    \centering
    \includegraphics[scale=0.25]{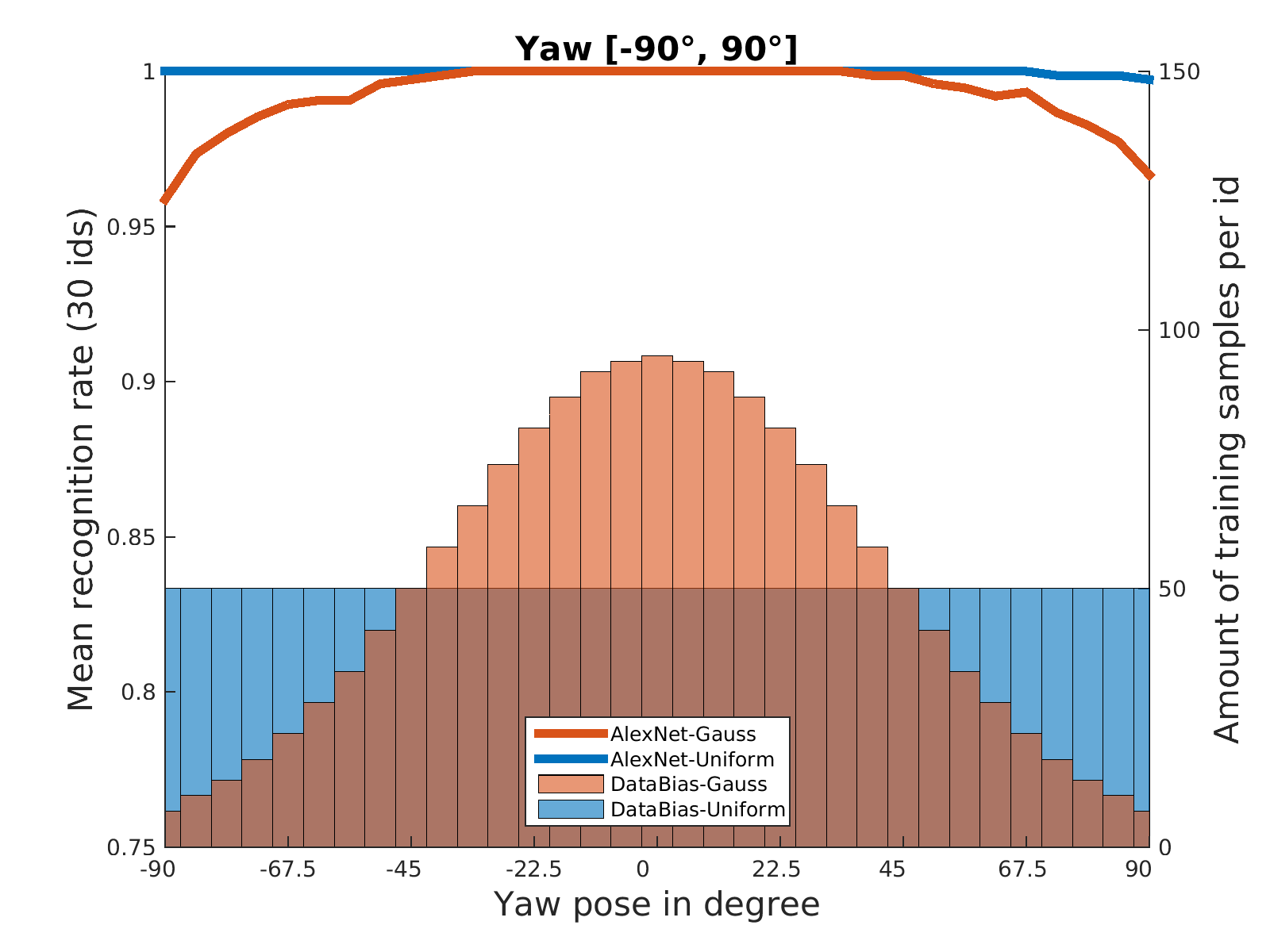}   
    \caption{Effect of biasing the training data to frontal faces. The plot shows the recognition rates of two AlexNet DCNNs as a function of the yaw pose. Both networks were trained on the same amount of images, however, the number of training samples per yaw pose is different. \textit{Blue curve:} TRR: $99.98\%$; Each yaw pose is equally likely to occur. \textit{Red curve:} TRR: $99.23\%$; Yaw pose is sampled according to a Gaussian distribution $\mathcal{N}(\mu=0^\circ,\sigma=7)$. The unbiased DCNN (blue) generalizes well along the axis of yaw variation, whereas the recognition rate of the biased DCNN drops significantly for those poses that are underrepresented in the training data.}
    \label{fig:DatasetBiasNonUniformSampling}
\end{figure}

\textbf{EXP-2: Non-uniform sampling of the yaw pose.} In Figure~\ref{fig:DatasetBiasNonUniformSampling} we illustrate the effect of biasing the yaw pose in the training data to frontal poses on the recognition performance at test time. Such non-uniform pose distributions are common in modern in the wild databases such as ALFW or Megaface.
The baseline curve in blue shows that a close to perfect recognition performance across the full yaw pose can be achieved, if the yaw pose is uniformly sampled in the training data. However, if a DCNN is trained on the same amount of training data but with a strong bias towards frontal poses then the recognition rate for faces in extreme poses drops significantly (red curve). Thus, we can deduce that an important property for face datasets is that the full variability of the yaw pose is reflected with a sufficient number of examples. In the supplementary, we show that the same generalization pattern can be observed for the VGG-16 architecture.

\begin{figure}
    \centering
    \subcaptionbox{\label{fig:DatasetBiasPlotSparse-88}}{\includegraphics[scale=0.25]{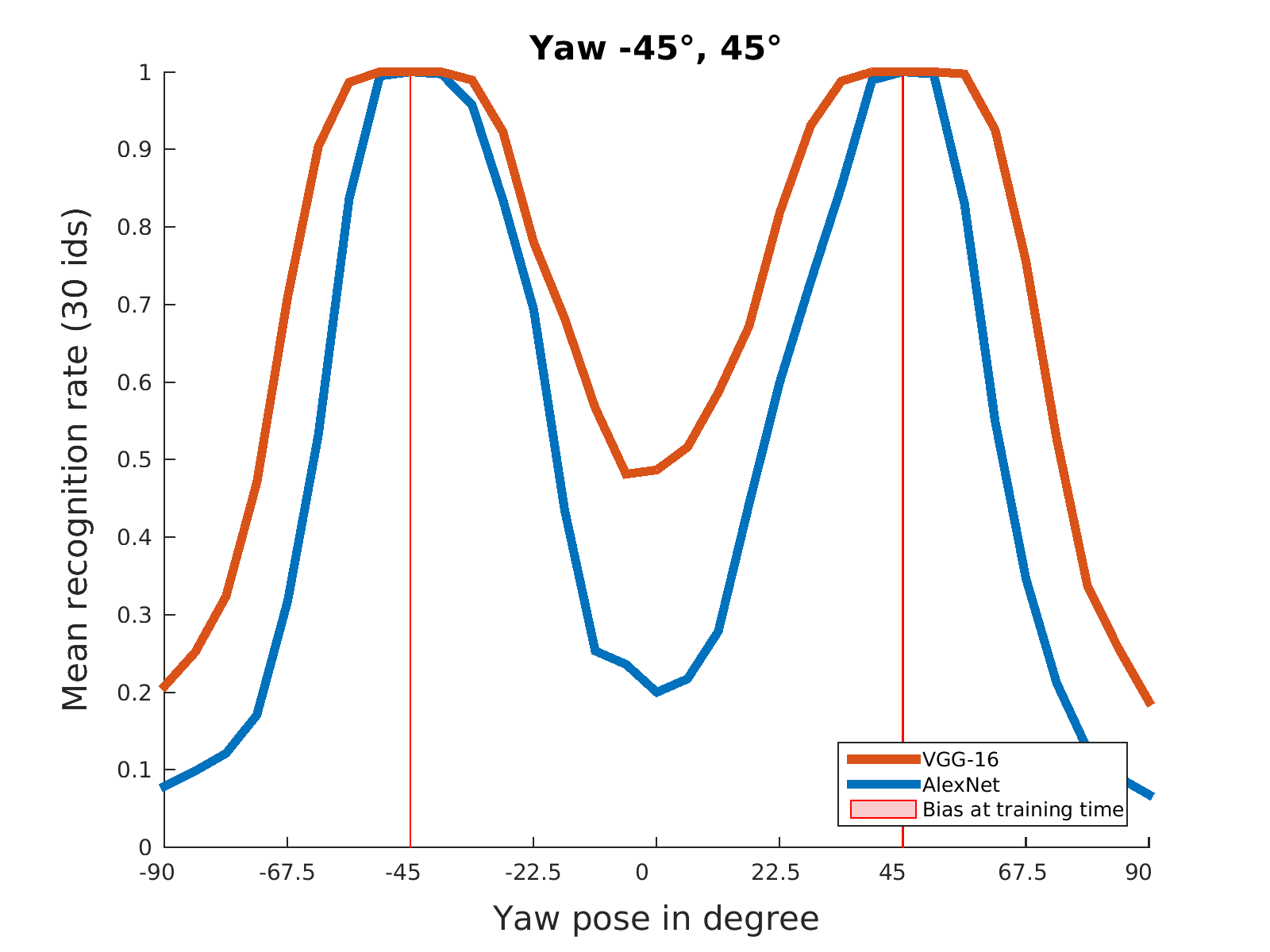}}
    \subcaptionbox{\label{fig:DatasetBiasPlotSparse-D8}}{\includegraphics[scale=0.25]{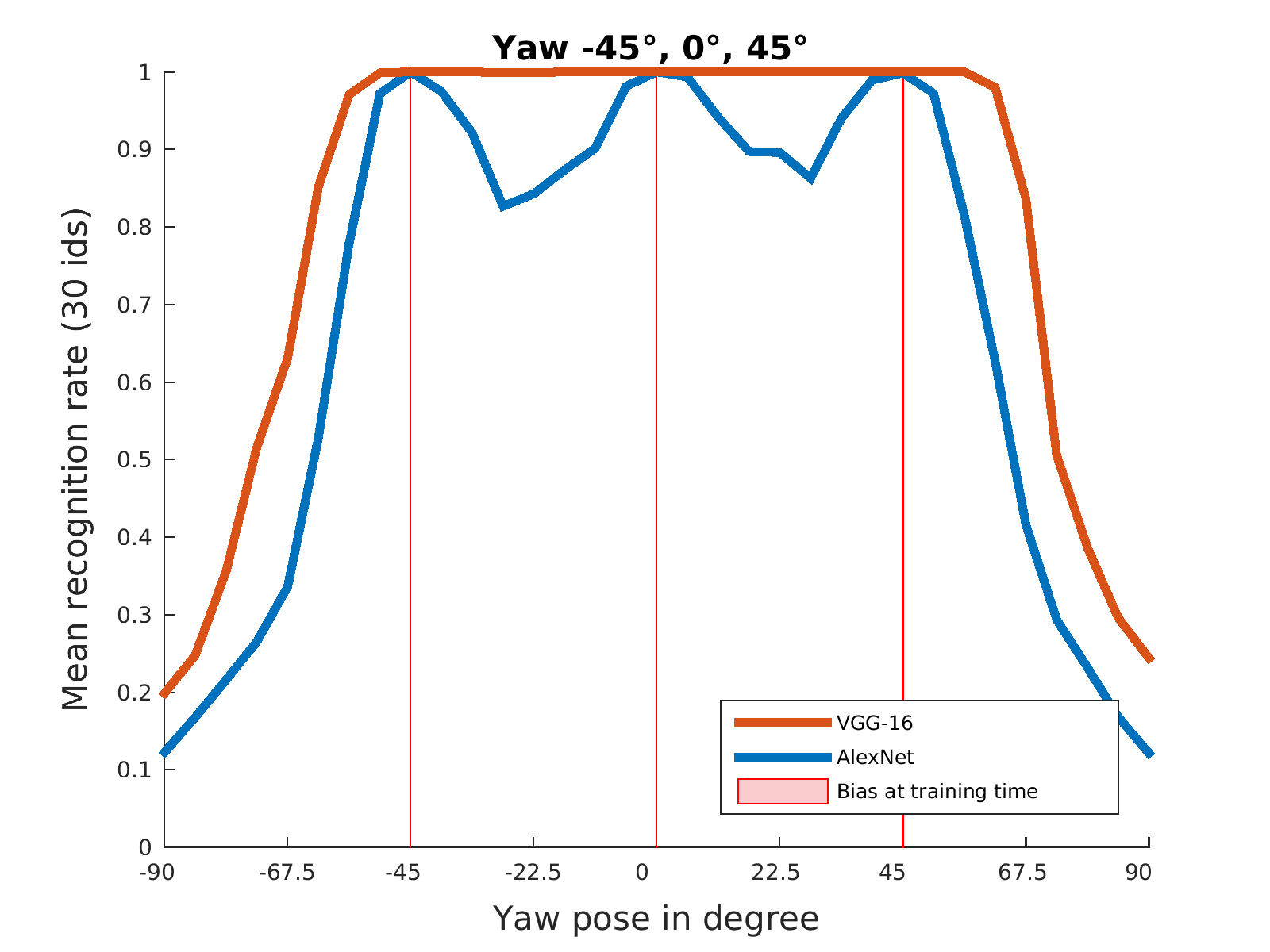}}
    \caption{Effect of sparsely sampling the yaw pose of faces at training time. (a)Yaw pose sampled at $-45^\circ$ and $45^\circ$ (AlexNet TRR: $51.8\%$; VGG-16 TRR: $70.5\%$); VGG-16 generalizes much better to frontal poses than AlexNet. (b) Yaw pose sampled at $-45^\circ$, $0^\circ$ and $45^\circ$ (AlexNet TRR: $69.3\%$; VGG-16 TRR: $81.9\%$); VGG-16 generalizes perfectly across the full range $[-45^\circ,45^\circ]$, whereas AlexNet still cannot generalize in between the sampled poses. }
    \label{fig:DatasetBiasPlotSparse}
\end{figure}

\textbf{EXP-3: Sparse sampling of the yaw pose.} In Figure~\ref{fig:DatasetBiasPlotSparse} we illustrate the effect of sampling the training data more sparsely along the axis of the yaw pose. We first bias the training set to yaw poses of $-45^\circ$ and $45^\circ$. VGG-16 achieves a TRR of $70.5\%$ at test time, whereas AlexNet only achieves $51.8\%$. Figure~\ref{fig:DatasetBiasPlotSparse-88} illustrates how these TRRs decompose as a function of the yaw pose. VGG-16 achieves constantly higher recognition rates across all poses. Most significantly, it is more than twice as good as AlexNet at recognizing frontal faces. If we add frontal faces at training time (Figure~\ref{fig:DatasetBiasPlotSparse-D8}) VGG-16 achieves a TRR of $81.9\%$, whereas AlexNet achieves $69.3\%$. Remarkably, VGG-16 is now able to recognize all faces correctly across the full range of $[-45^\circ,45^\circ]$, whereas the recognition rates of AlexNet still drop significantly for poses in between $[-45^\circ,0^\circ]$ and $[0^\circ,45^\circ]$. Thus, the architecture of VGG-16 enables the DCNN to generalize well from only a few well distributed example views to other unseen views, although it has more parameters than AlexNet.

\begin{figure}
    \centering
    \includegraphics[scale=0.25]{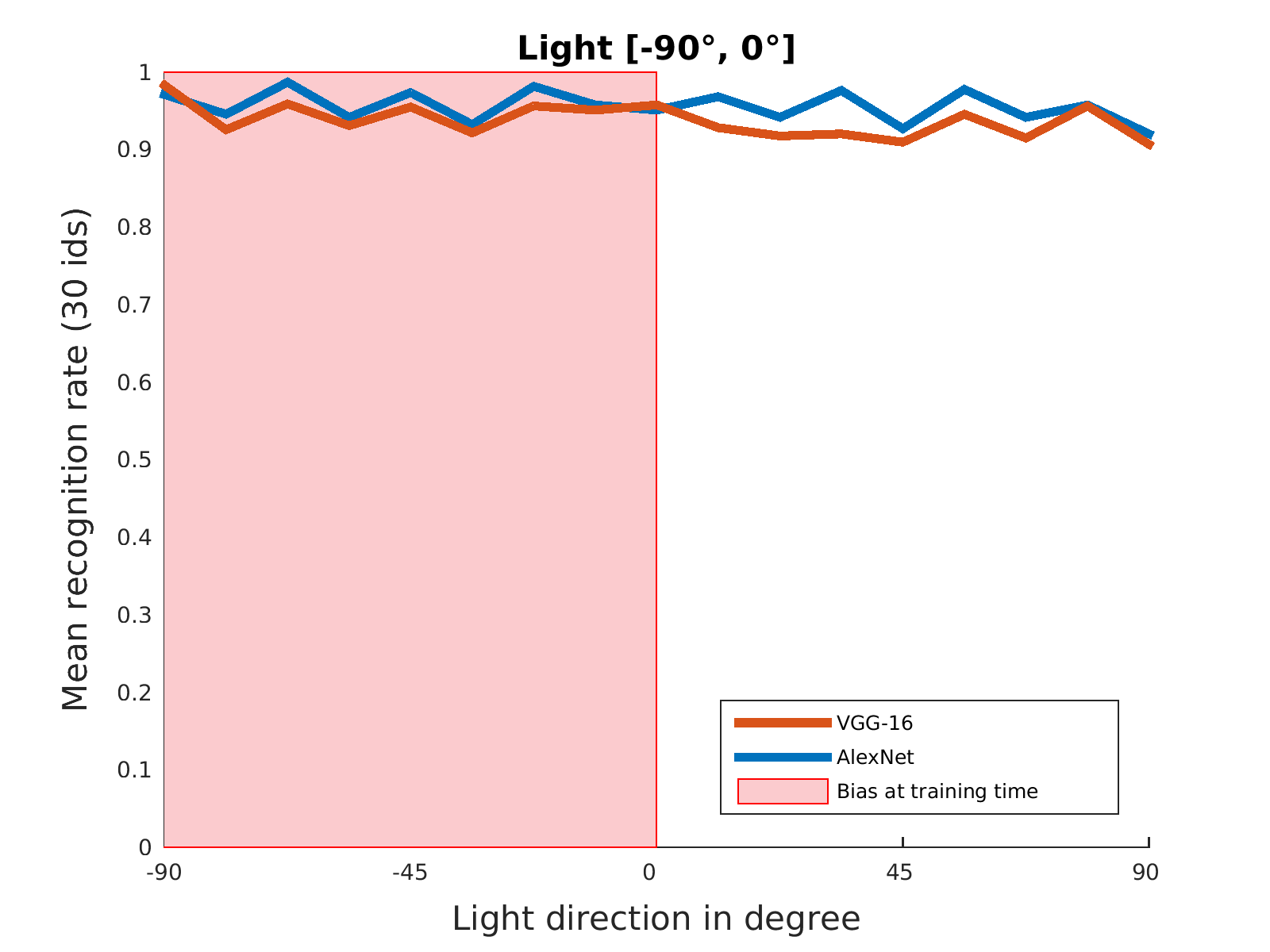}
    \caption{Effect of biasing the light direction at training time. In the experiment the pose of all faces is fixed to be frontal. Face images in the training vary in terms of light direction in the range $[-90^\circ,0^\circ]$. At test time, AlexNet and VGG-16 generalize well to the unseen illumination conditions $>0^\circ$ (AlexNet TRR: $95.6\%$; VGG-16 TRR: $93.7\%$).}
    \label{fit:DatasetBiasLight}
\end{figure}

\textbf{EXP-4: Bias in the illumination.} In this experiment, we test how strong the effect of a bias in the illumination condition is on the recognition performance. We fix the pose of faces in the training data to be frontal and only vary the light direction. We restrict the variation in the light direction at training time to the range $[-90^\circ,0^\circ]$. Figure~\ref{fit:DatasetBiasLight} illustrates, that both DCNN types can generalize very well to the unseen illumination conditions. This might be due to the fact that our illumination model does not include self-shadowing and hard shadows. Thus by focusing on the image gradient information a DCNN could strongly limit the influence of changing illumination conditions. 


\begin{figure}
    \centering
    \includegraphics[scale=0.25]{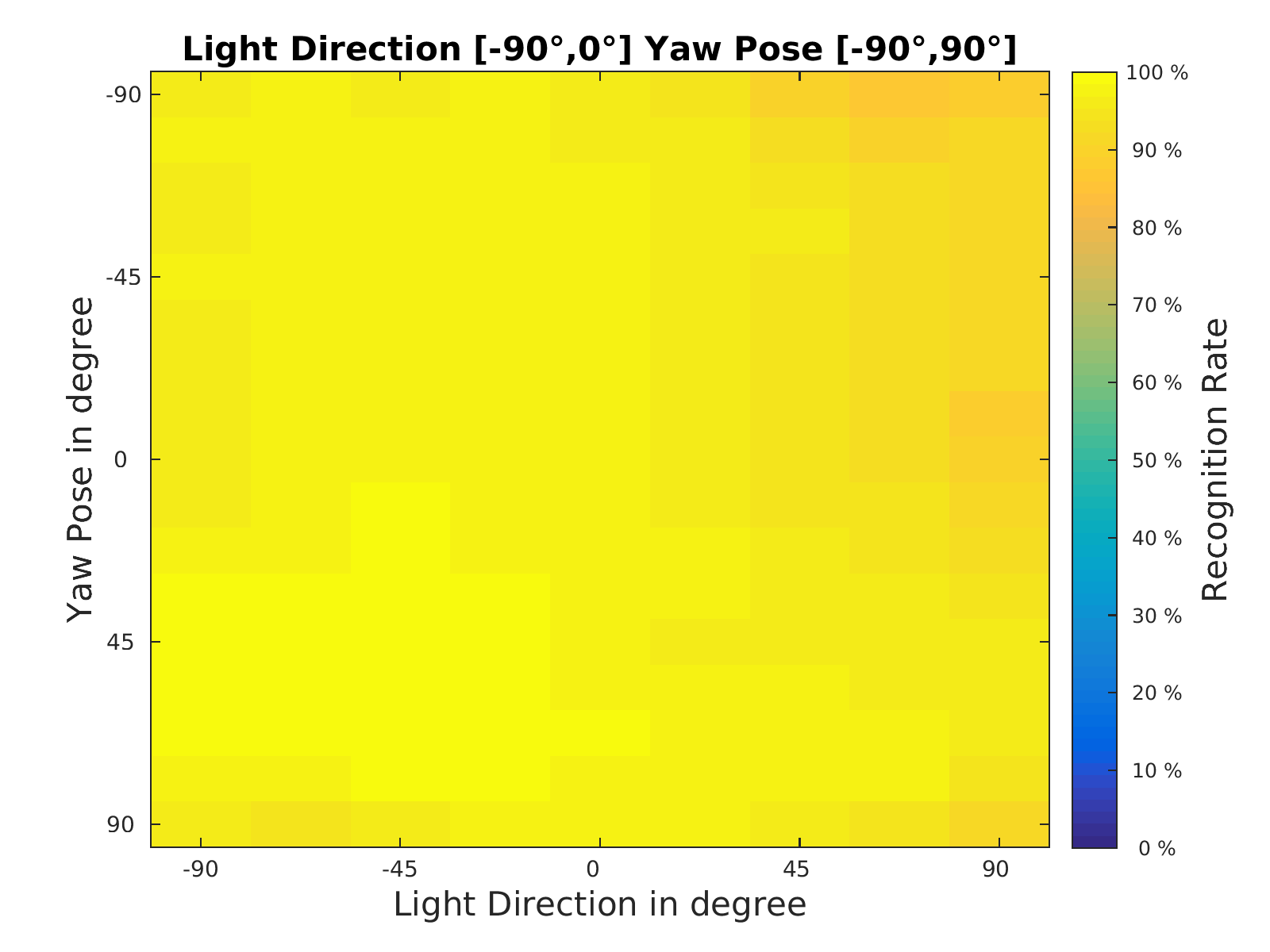}
    \caption{Illustration of the recognition rate as a function of the light direction and yaw pose for a DCNN with AlexNet architecture. The light direction in the training data was biased to the range $[-90^\circ,0^\circ]$, while the yaw pose varied in the full range $[-90^\circ,90^\circ]$. The DCNN can generalize well even to previously unseen combinations of yaw pose and light direction.}
    \label{fig:poselight}
\end{figure}

\textbf{EXP-5: Bias in the illumination with pose variation.} In the following experiment, we test if an AlexNet DCNN can still generalize under biased illumination conditions when the face pose is variable. In particular, we vary faces in the training set uniformly across the full yaw range $[-90^\circ,90^\circ]$. As in the previous experiment EXP-4, we restrict the variation of the light direction to the range $[-90^\circ,0^\circ]$. 
Figure~\ref{fig:poselight} illustrates the recognition rate as a function of the yaw pose and light direction.
We can clearly observe that the DCNN generalizes well across the full pose variation and across the full range of light direction. This is surprising because the effect of the pose-light interaction on the facial appearance has not been observed at training time for light directions $>0^\circ$. We think that the DCNN can generalize to unseen light directions very well, because these transformations only have a relatively small impact on the gradients in the images compared to changes in the identity or variations in the pose. Therefore, we suppose that DCNNs trained on face recognition might have a strong focus on gradient information in the image.

\textbf{Summary.} In this section, we have shown that in order to achieve a good face recognition performance across the yaw pose, the full pose variation must be reflected in the training data with a sufficient number of training samples. However, the parameters of the yaw pose must not be sampled densely when training with the VGG-16 architecture  (Fig.\ref{fig:DatasetBiasPlotSparse}). Furthermore, we have observed that DCNNs can generalize surprisingly well to unseen facial appearances due to changing light directions.
In all experiments with missing viewpoints, we have seen that the DCNNs with the VGG-16 architecture can significantly better generalize than DCNNs with the AlexNet architecture.

\begin{figure}
    \centering
    \subcaptionbox{\label{fig:dis1-a}}{\includegraphics[scale=0.25]{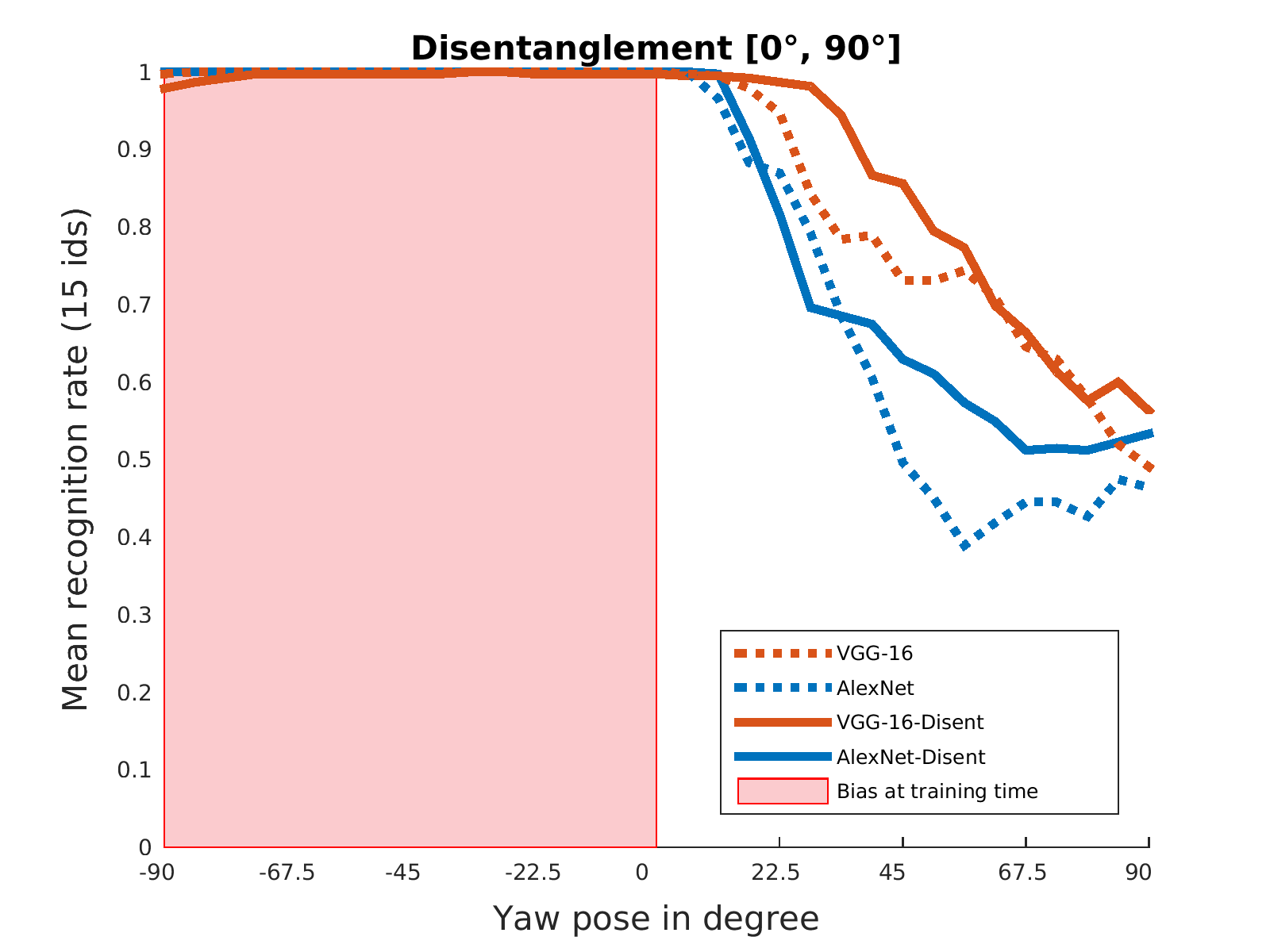}}
    \subcaptionbox{\label{fig:dis1-b}}{\includegraphics[scale=0.25]{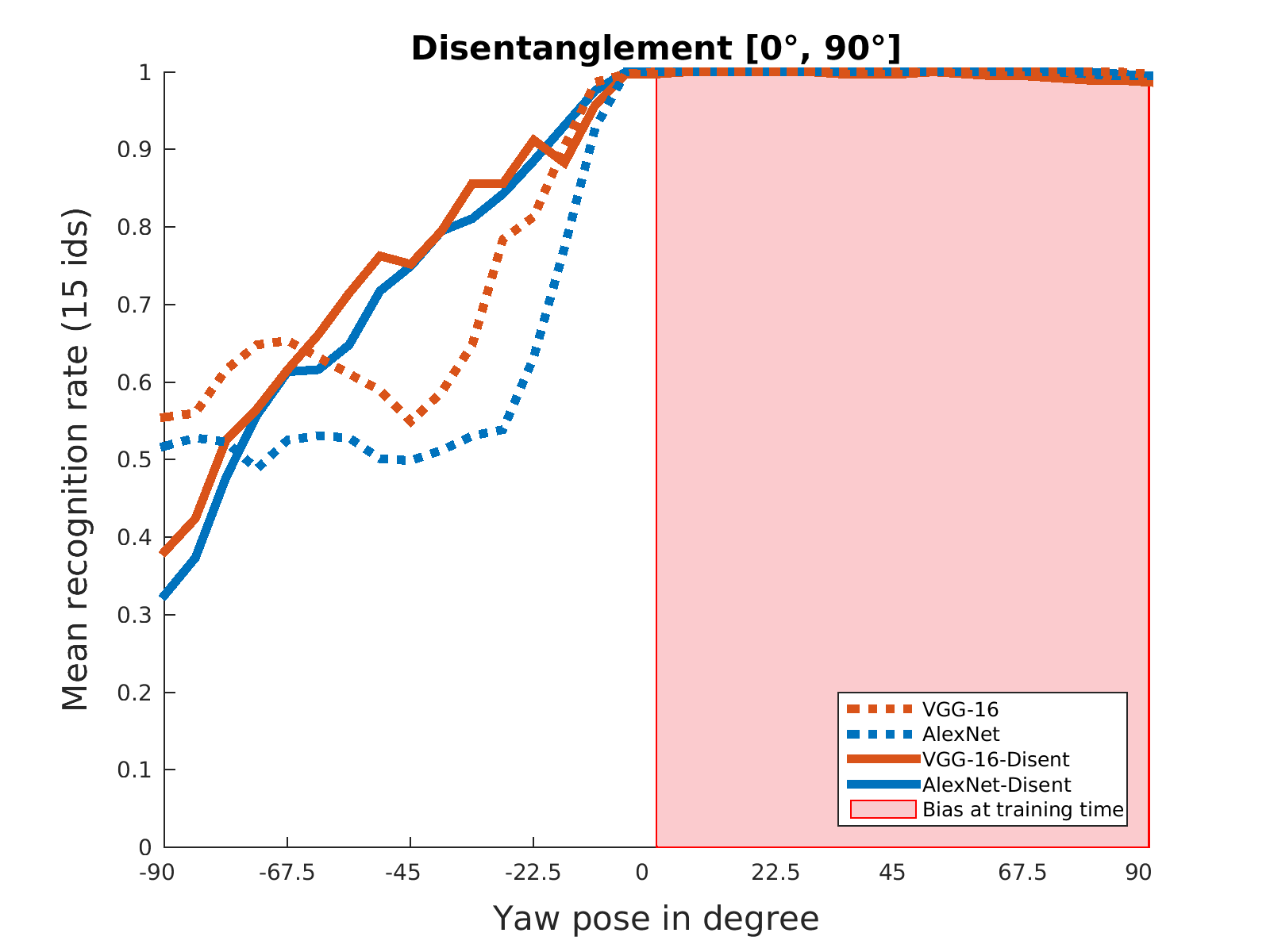}}
    \caption{ Testing disentanglement ability of DCNNs. Dotted lines: DCNNs trained on a biased yaw pose (illustrated in Figure~\ref{fig:SetupDisentanglement-partial}). Solid lines: Disentanglement setup (illustrated in Figure~\ref{fig:SetupDisentanglement-full}). (a) Left-Identities with biased yaw pose of [$-90^\circ, 0^\circ$]. (b) Right-Identities with biased yaw pose of [$0^\circ, 90^\circ$]. DCNNs cannot make use of the additional information about the pose transformation which is present in the data in the disentanglement setup.}
    \label{fig:dis1}
\end{figure}
\subsection{Disentanglement bias across facial identities}
\label{sec:experiments-disentanglement}
In the previous section, we have observed that DCNNs generalize well as soon as a nuisance transformation is sufficiently represented for \textit{each} identity in the training. When this was not the case, the generalization performance decreased significantly. In this section, we study if DCNNs are capable of generalizing if the nuisance transformation is densely reflected in the training data across multiple identities. 
In particular, each face identity in the training is varied in a certain interval of the yaw pose. However, across all identities the full yaw pose variation is reflected. In Figure~\ref{fig:SetupDisentanglement-full} we schematically illustrate how this setup compares to the one from the previous Section~\ref{sec:experiments-bias} (Figure~\ref{fig:SetupDisentanglement-partial}). We call this type of bias \textit{disentanglement bias}, since if DCNNs are capable of disentangling the image variation induced by the yaw pose from the face identity, then they would be able to generalize well on this dataset.

\textbf{EXP-6: Disentanglement of pose variation.} In this experiment, half of the identities in the training set vary in the yaw pose range of [$-90^\circ, 0^\circ$]. We refer to those identities as the set Left-identities. The other half of the faces varies in the range $[0^\circ, 90^\circ]$ (Right-identities, Figure~\ref{fig:SetupDisentanglement-full}). 
Figure~\ref{fig:dis1} illustrates the recognition performance of DCNNs trained on the full training set. We evaluate the Left-identities and Right-identities separately (Figure~\ref{fig:dis1-a} \& Figure~\ref{fig:dis1-b}).
We observe, that the DCNNs only slightly improve compared to setup where the yaw pose range is restricted to [$-90^\circ, 0^\circ$] for all identities (dotted curves). Thus, both DCNNs cannot benefit from the additional information in the training set.
We conclude that this phenomenon occurs because they are not able to disentangle the image variation induced by the pose variation and the identity change. 
\begin{figure}
    \centering
    \subcaptionbox{\label{fig:DisentanglementRegularization-Left}}{\includegraphics[scale=0.25]{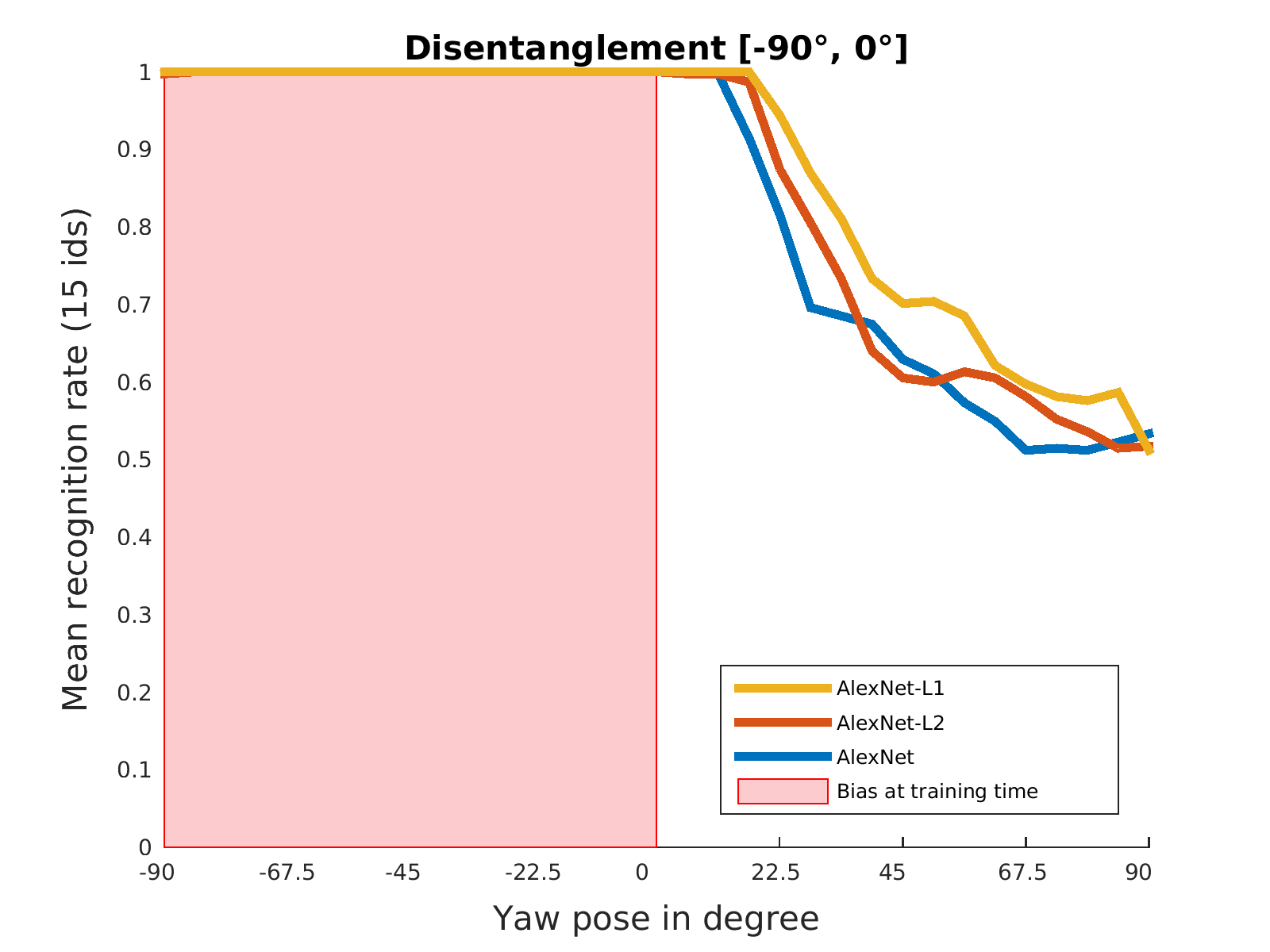}}
    \subcaptionbox{\label{fig:DisentanglementRegularization-Right}}{\includegraphics[scale=0.25]{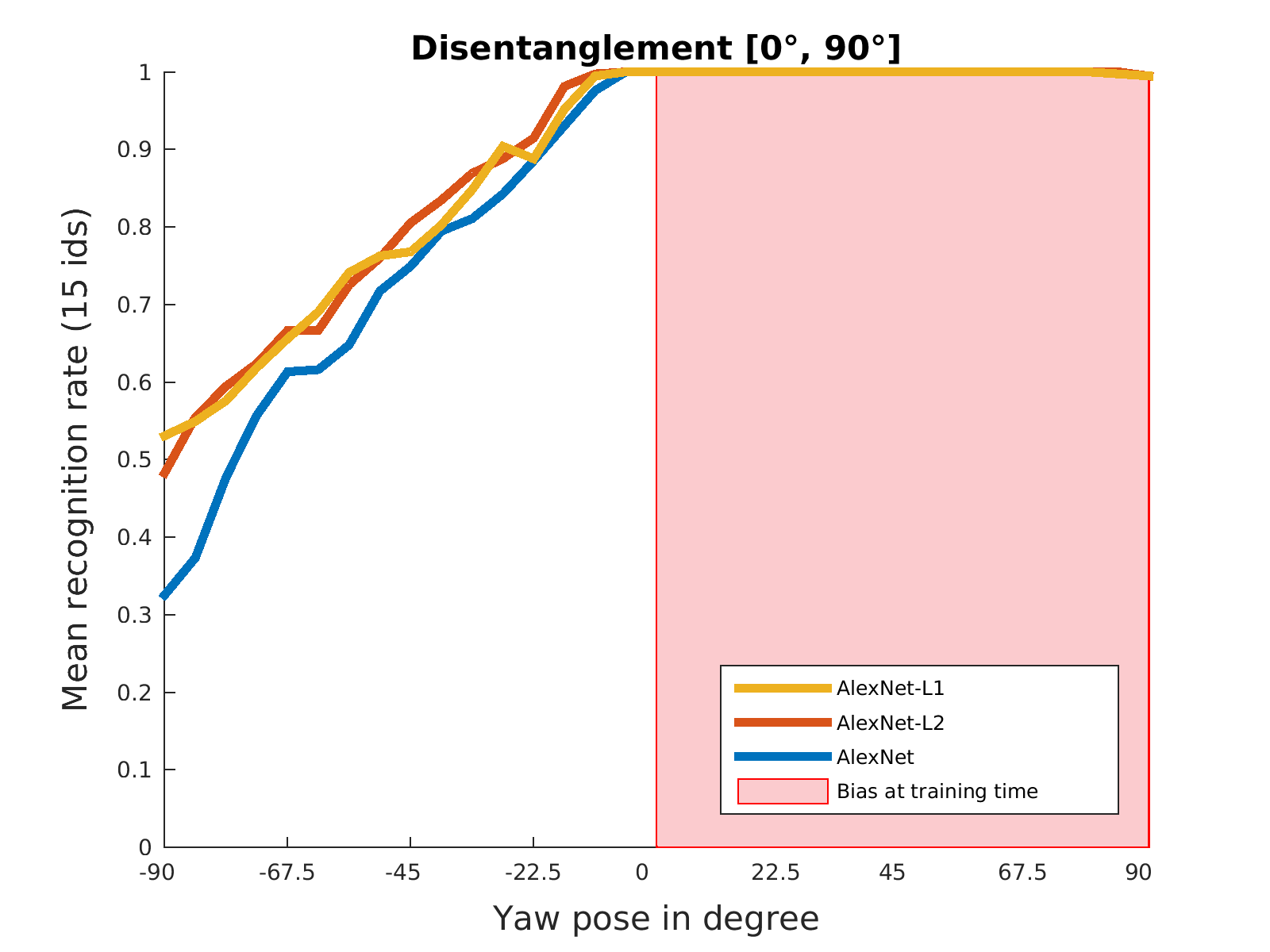}}
    \caption{Influence of regularization on the ability of AlexNet to disentangle identity and pose transformation. (a) Left-Identities. (b) Right-Identities. Strongly regularizing AlexNet with $L_1$ (yellow) or $L_2$ regularization on the weights, does slightly improve the networks disentanglement ability, compared to a weak regularization (blue). }
    \label{fig:DisentanglementRegularization}
\end{figure}

\textbf{EXP-7: Influence of regularization on disentanglement ability.} We test if a strong regularization on the network weights improves the performance of DCNNs in the disentanglement setup. The hypothesis underlying this experiment is that the capacity of the network might be too large, which favors memorization of the training examples and hinders it from performing disentanglement. Therefore, we increase the weight decay parameter $\lambda$ during SGD. To find the strongest possible regularization, we increase $\lambda$ up to the point where the training of the networks does not succeed anymore and set $\lambda$ to be the penultimate value. We use the AlexNet architecture and apply regularization weights $\lambda_{L1}=0.001$ as well as $\lambda_{L2}=0.01$ (Figure~\ref{fig:DisentanglementRegularization}). A strong regularization does not significantly increase a DCNNs ability to perform disentanglement. In the supplementary material, we show that the same generalization patterns can be observed for the VGG-16 network.\\
\textbf{Summary.} We have observed that DCNNs which are trained from scratch are not able to disentangle the image variation induced by pose transformations from the one induced by the change of the identity. 
This suggests that DCNNs cannot perform disentanglement if the space of nuisance transformations is not reflected in the training samples of each identity in the training set. The proposed benchmark is perfectly suited to analyze the disentanglement performance of novel DCNN architectures.
\begin{figure}[t]
    \centering
    \subcaptionbox{\label{fig:ExperimentMultipie-1}}{\includegraphics[scale=0.20]{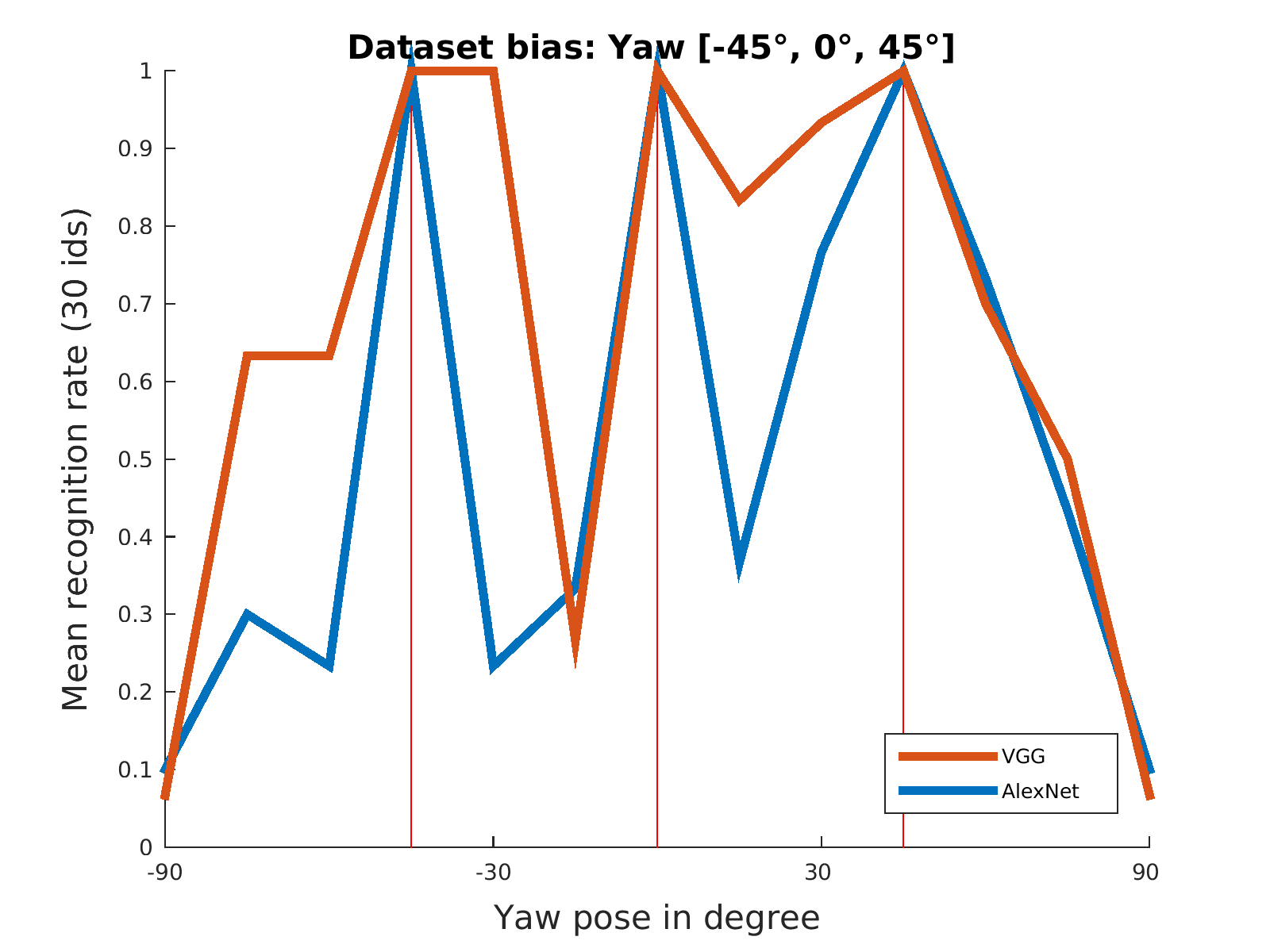}}
    \subcaptionbox{\label{fig:ExperimentMultipie-2}}{\includegraphics[scale=0.20]{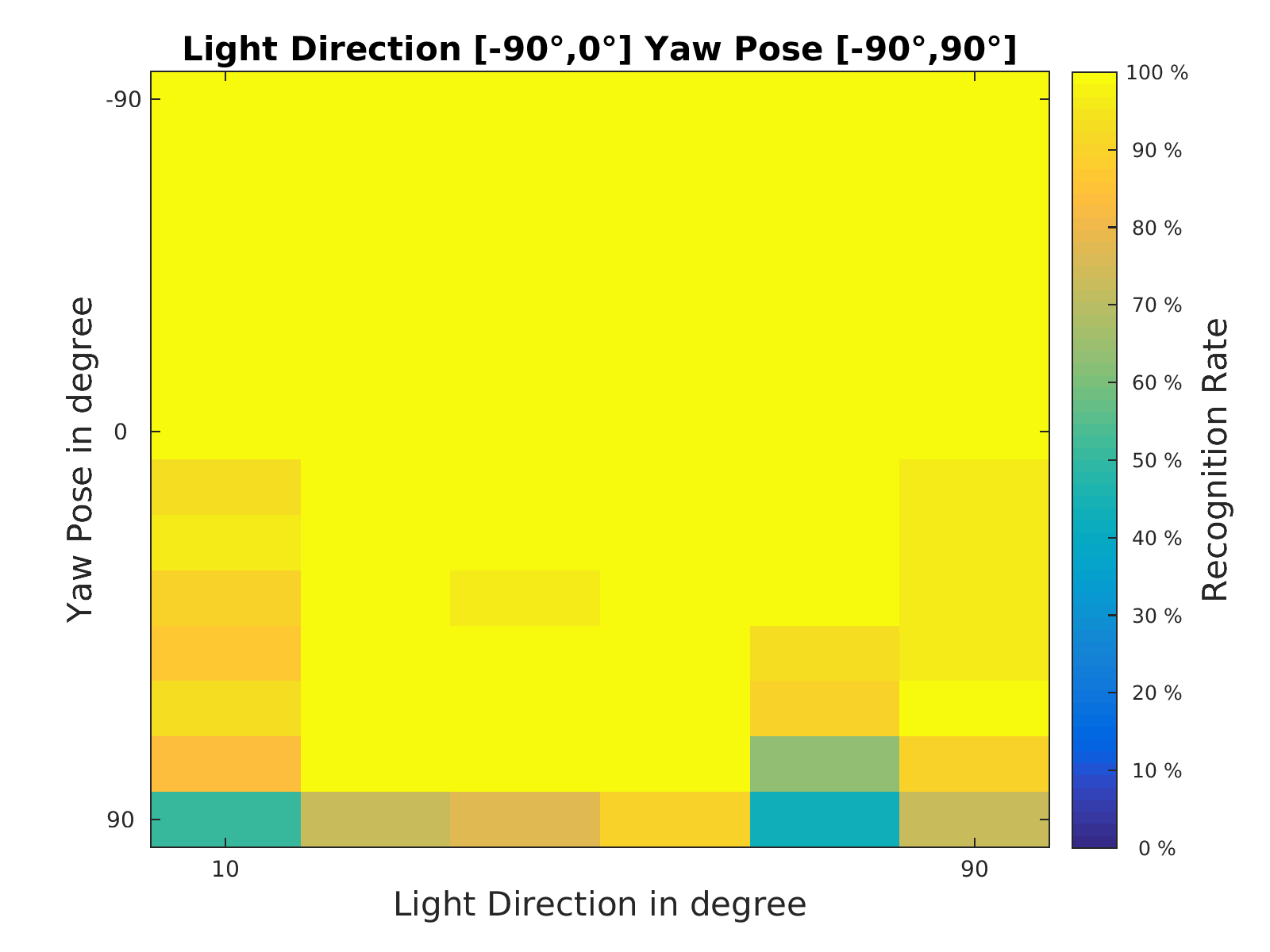}}
    \caption{Reproduction of the experiments EXP-3 and EXP-8 on real data (Figure~\ref{fig:DatasetBiasPlotSparse-D8} and \ref{fig:poselight} ). (a) Sparse sampling of the training data at yaw poses $-45^\circ$,$0^\circ$ and $45^\circ$. (b) Bias of the illumination direction to [$-90^\circ, 0^\circ$] and full yaw pose variation. In both cases, the generalization patterns are very similar to the ones obtained on the synthetic data.}
    \label{fig:ExperimentMultipie}
\end{figure}
\subsection{Validity for real data}
\label{sec:experiments-realdata}
In this section, we study if the generalization patterns that we observed on synthetic data can be reproduced on real data.
The CMU Multi-PIE \cite{gross2010multi} database is one of the biggest datasets with annotated facial pose and illumination conditions. Our experiments in this section should be regarded more as proof of the concept behind our methodology, rather than as evidence that all of our observations transfer one-to-one to real data. We use data of $30$ identities of session-01. For training, we crop the face in a $300\times300$ region and rescale it to have size $227\times227$. \\
\textbf{EXP-8: Real data - Sparse sampling of yaw pose.}
In this experiment, we reproduce the setup of experiment EXP-3. We use frontal illumination (flash  $16$) at training time. At test time, we classify the same identities in a slightly different illumination setup (flash $15$). Analogous to experiment EXP-3, we bias the yaw pose in the training data to the poses $-45^\circ$, $0^\circ$ and $45^\circ$. 
Figure~\ref{fig:ExperimentMultipie-1} illustrates that the generalization performance of both DCNNs is very similar to what we have observed on the synthetic data. Compared to AlexNet, VGG-16 generalizes much better in the yaw pose range of $[-45^\circ, 45^\circ]$. Beyond this range, the recognition performance of both networks drops significantly.\\
\textbf{EXP-9: Real data - Bias in the illumination with pose variation.}
We reproduce the setup of experiment EXP-5. At training time, we use the light directions in the range [$-90^\circ, 0^\circ$] (flash $0-6$) for the full pose range [$-90^\circ, 90^\circ$]. At test time, we only classify faces with light coming from the directions [$0^\circ, 90^\circ$] (flash numbers $7-13$). We train the AlexNet architecture and illustrate the results in Figure~\ref{fig:ExperimentMultipie-2}. Again, the generalization pattern is very similar to the one observed on synthetic data. The DCNN can generalize very well to unseen illuminations.\\
In summary, we observed that the generalization patterns from experiments EXP-3 and EXP-5 on synthetic data can also be observed when training on real world data.
\section{Conclusion}
\label{sec:conclusion}
In this work, we have studied the effect of dataset bias and DCNN architectures on the generalization performance of deep face recognition systems with a fully parametric generator of face images. 
We demonstrated that the full control over the image variation makes possible to decompose the recognition score as a function of nuisance transformations. This enabled us to systematically analyze and compare DCNNs at the task of face recognition. \\
We verified that biases in the pose distribution have a significant influence on the generalization performance 
while this is not the case for biases in the illumination.\\
We used the proposed methodology to study \textit{why} the VGG-16 architecture generally outperforms the AlexNet architecture at face recognition tasks. We showed that a major reason for this phenomenon is that VGG-16 can better generalize from missing data in the pose distribution as well as from a bias to frontal face poses.\\
A major limitation of the analyzed DCNN architectures is that they have severe difficulties to generalize when different identities do not share the same pose variation. Lastly, we collected evidence that the generalization patterns we observe when training on synthetic data, also occur when training on real data. Our findings have to be taken with some caveats. Our training setups were controlled and have to be confirmed on larger datasets with millions of identities and additional combinations of nuisance transformations. Nevertheless, our findings raise fundamental questions about the generalization patterns that we observed: 1) What is the mechanism that allows VGG-16 better generalize to large unseen poses? 2) Why can DCNNs generalize so well to unseen illumination conditions, although they have a significant effect on the facial appearance? 3) What additional mechanisms would lead to a better disentanglement of pose variations across identities?\\
Our face image generator is publicly available and allows to compare DCNN architectures on a common ground, as well as to understand their internal mechanisms better.
{\small
\bibliographystyle{ieee}
\bibliography{egbib}
}   
\newpage
\textcolor{white}{.}
\includepdf[scale=1,pages=1]{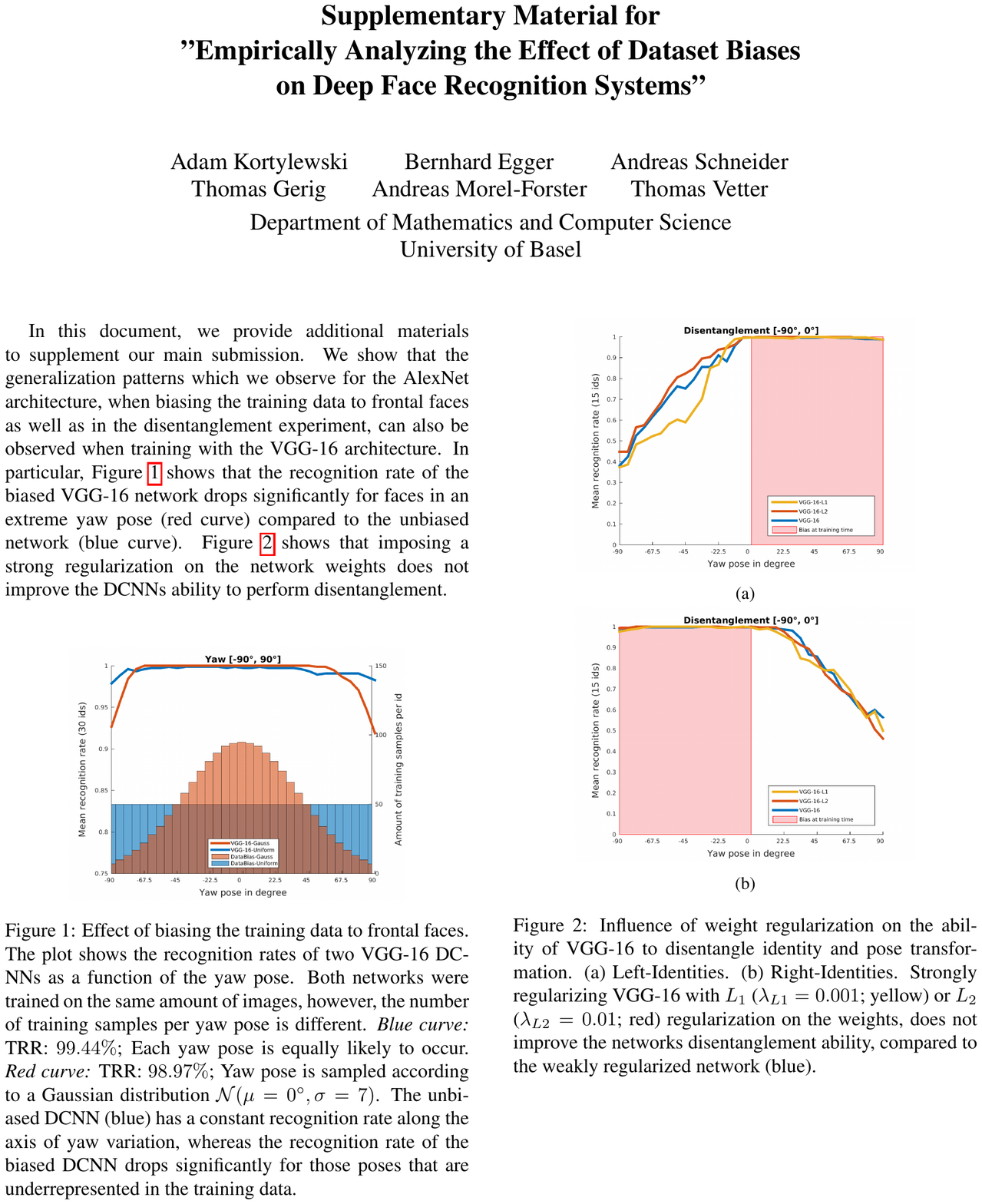}

\end{document}